\documentclass[twocolumn]{paper}
\usepackage[sort&compress,square,comma,authoryear]{natbib}
\usepackage{float}
\usepackage[affil-it]{authblk}
\usepackage{amsmath}
\usepackage{mathtools}
\usepackage{graphicx}
\usepackage{amssymb}
\usepackage{lineno}

\usepackage{commath}

\begin{document}

%\begin{frontmatter}

%% Title, authors and addresses

\title{\vspace{-3.0cm}STDP allows close-to-optimal spatiotemporal spike pattern detection by single coincidence detector neurons}

%% use the tnoteref command within \title for footnotes;
%% use the tnotetext command for the associated footnote;
%% use the fnref command within \author or \address for footnotes;
%% use the fntext command for the associated footnote;
%% use the corref command within \author for corresponding author footnotes;
%% use the cortext command for the associated footnote;
%% use the ead command for the email address,
%% and the form \ead[url] for the home page:
%%
%% \title{Title\tnoteref{label1}}
%% \tnotetext[label1]{}
%% \author{Name\corref{cor1}\fnref{label2}}
%% \ead{email address}
%% \ead[url]{home page}
%% \fntext[label2]{}
%% \cortext[cor1]{}
%% \address{Address\fnref{label3}}
%% \fntext[label3]{}

%% use optional labels to link authors explicitly to addresses:
%% \author[label1,label2]{<author name>}
%% \address[label1]{<address>}
%% \address[label2]{<address>}

%\author{Timoth\'ee Masquelier }
%\ead{timothee.masquelier@cnrs.fr}

%\affil{CERCO UMR5549 CNRS - Universit\'e Toulouse 3, France}

\author{Timoth\'ee Masquelier%
  \thanks{e-mail: \texttt{timothee.masquelier@cnrs.fr}}}
\affil{CERCO UMR5549 CNRS - Universit\'e Toulouse 3, France}

\maketitle

\begin{abstract}
%% Text of abstract
%%maximum length 300 words
By recording multiple cells simultaneously, electrophysiologists have found evidence for repeating spatiotemporal spike patterns. In sensory systems in particular, repeating a sensory sequence typically elicits a reproducible spike pattern, which carries information about the sensory sequence. How this information is extracted by downstream neurons is unclear. In this theoretical paper, we investigate to what extent a single cell could detect a given spike pattern and what the optimal parameters to do so are, in particular the membrane time constant $\tau$. Using a leaky integrate-and-fire (LIF) neuron with instantaneous synapses and homogeneous Poisson input, we were able to compute this optimum analytically. Our results indicate that a relatively small $\tau$ (at most a few tens of ms) is usually optimal, even when the pattern is much longer. This is somewhat counter intuitive as the resulting detector ignores most of the pattern, due to its fast memory decay. Next, we wondered if spike-timing-dependent plasticity (STDP) could enable a neuron to reach the theoretical optimum. We simulated a LIF neuron equipped with additive spike-timing-dependent potentiation and homeostatic rate-based depression, and repeatedly exposed it to a given input spike pattern. As in previous studies, the LIF progressively became selective to the repeating pattern with no supervision, even when the pattern was embedded in Poisson activity. Here we show that, using certain STDP parameters, the resulting pattern detector can be optimal. Taken together, these results may explain how humans can learn repeating visual or auditory sequences. Long sequences could be recognized thanks to coincidence detectors working at a much shorter timescale. This is consistent with the fact that recognition is still possible if a sound sequence is compressed, played backward, or scrambled using 10ms bins. Coincidence detection is a simple yet powerful mechanism, which could be the main function of neurons in the brain.

\end{abstract}

%\begin{keyword}
\noindent\textbf{Keywords:}\\
Spike-timing-dependent plasticity (STDP), leaky integrate-and-fire neuron, coincidence detection, multi-neuron spike sequence, spatiotemporal spike pattern, unsupervised learning
%% keywords here, in the form: keyword \sep keyword

%% MSC codes here, in the form: \MSC code \sep code
%% or \MSC[2008] code \sep code (2000 is the default)

%\end{keyword}

%\end{frontmatter}

%%
%% Start line numbering here if you want
%%
% \linenumbers

%% main text
\section{Introduction}

Electrophysiologists report the existence of repeating spike sequence involving multiple cells, also called ``spatiotemporal spike patterns'', with precision in the millisecond range, both \emph{in vitro} and \emph{in vivo}, lasting from a few tens of ms to several seconds \citep{Tiesinga2008}. In sensory systems, different  stimuli evoke different spike patterns (also called ``packets'') \citep{Luczak2015}. In other words, the spike patterns contain information about the stimulus. How this information is extracted by downstream neurons is unclear. Can it be done by neurons only one synapse away from the recorded neurons? Or are multiple integration steps needed? Can it be done by simple coincidence detector neurons, or should other temporal features, such as spike ranks, be taken into account? Here we wondered how far we can go with the simplest scenario: the readout is done by simple coincidence detector neurons only one synapse away from the neurons involved in the repeating pattern. We demonstrate that this approach can lead to very robust pattern detectors, provided that the membrane time constants are relatively short, possibly much shorter than the pattern duration.

%In this theoretical paper, we first investigate to what extent single neurons can detect the presence of a given spike pattern, and what are the optimal parameters to do so. Using a leaky integrate-and-fire neuron (LIF) with instantaneous synapses and homogeneous Poisson inputs, we were able to compute this optimum analytically.

In addition, it is known that mere repeated exposure to meaningless sensory sequences facilitates their recognition afterwards, in the visual \citep{Gold2014} and auditory modalities \citep{Agus2010,Andrillon2015,Viswanathan2016}, even when the subjects were unaware of these repetitions. Thus, an unsupervised learning mechanism must be at work. It could be the so called spike-timing-dependent plasticity (STDP). Indeed, some theoretical studies by us and others have shown that neurons equipped with STDP can become selective to arbitrary repeating spike patterns, even without supervision~\citep{Masquelier2008,Masquelier2009,Gilson2011,Humble2012,Hunzinger2012,Klampfl2013,Nessler2013,Kasabov2013a,Krunglevicius2015a,Sun2016,Yger2015}. Using numerical simulations, we show here that the resulting detectors can be close to the theoretical optimum.

\section{Formal description of the problem}

\begin{figure}[H]
\centering\includegraphics[width=0.9\linewidth]{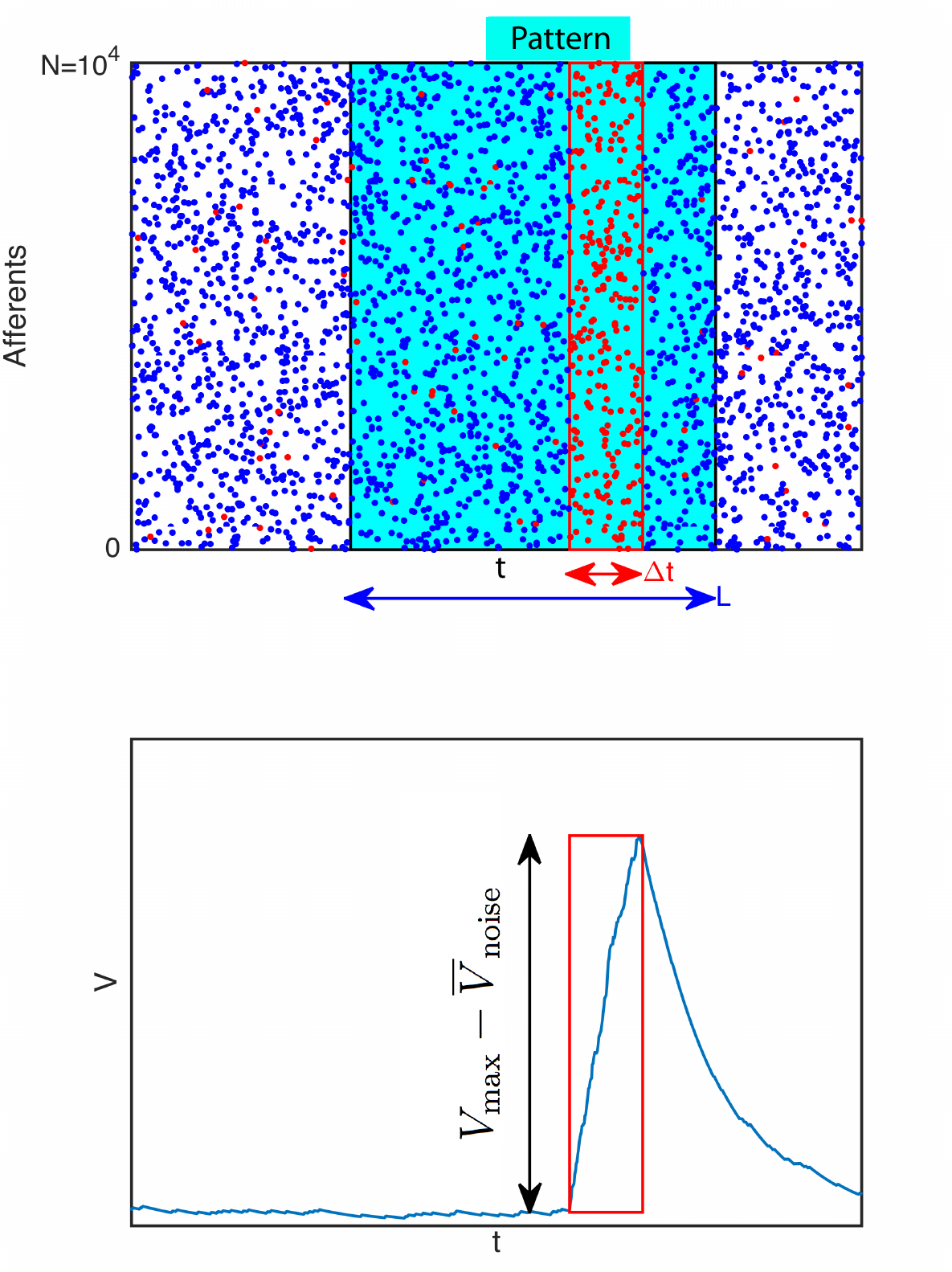}
\caption{Detecting a spike pattern with a LIF neuron. (Top) Raster plot of $N=10^4$ neurons firing according to an homogeneous Poisson process. A pattern of duration $L$ can be repeated (frozen noise). Here we illustrated Strategy~\#1, which consists in connecting the LIF to all neurons that fire at least once during a certain time window of the pattern, with duration $\Delta t\leq L$. These neurons emit red spikes. Of course they also fire outside of the $\Delta t$ window. (Bottom) Typically the LIF's potential will be particularly high when integrating the spikes of the $\Delta t$ window, much higher than with random Poisson inputs, and we want to optimize this difference, or more precisely the signal-to-noise ratio (SNR, see text).}
\label{fig:strat}
\end{figure}

We assess the problem of detecting a spatiotemporal spike pattern with a single LIF neuron. Intuitively, one should connect the LIF to the neurons that are particularly active during the pattern, or during a subsection of it. That way, the LIF will tend to be more activated by the pattern than by some other input. More formally, we note $L$ the pattern duration, $N$ the number of neurons it involves. We call Strategy~$\#n$ the strategy which consists in connecting the LIF to the $M$ neurons that emit at least $n$ spike(s) during a certain time window $\Delta t \le L$ of the pattern. Strategy~\#1 is illustrated on Figure~\ref{fig:strat}.

We hypothesize that all afferent neurons fire according to an homogeneous Poisson process with rate $f$, both inside and outside the pattern. That is the pattern corresponds to one realization of the Poisson process, which can be repeated (this is sometimes referred to a ``frozen noise''). To model jitter, at each repetition a random time lag is added to each spike, drawn from a uniform distribution over $[-T,T]$ (a normal distribution is more often used, but it would not allow analytical treatment, see next section).

We also assume that synapses are instantaneous (i.e. excitatory postsynaptic currents are made of Diracs), which facilitates the analytic calculations.

For now we ignore the LIF threshold, and we want to optimize its signal-to-noise ratio (SNR), defined as:
\begin{equation}
SNR =  \frac{V_{\mathrm{max}}-\overline{V}_{\mathrm{noise}}} {\sigma_{\mathrm{noise}}},
\end{equation}
where $V_{\mathrm{max}}$ is the maximal potential reached during the pattern presentation,  $\overline{V}_{\mathrm{noise}}$ is the mean value for the potential with Poisson input (noise period), and $\sigma_{\mathrm{noise}}$ its standard deviation (see Figure  \ref{fig:strat}).

\section{A theoretical optimum}
\subsection{Deriving the SNR analytically}
We now want to calculate the SNR analytically. In this section, we assume unitary synaptic weights. Since the LIF has instantaneous synapses, and the input spikes are generated with a Poisson process, we have
$\overline{V}_{\mathrm{noise}} = \tau f M$ and
$\sigma_{\mathrm{noise}} = \sqrt{\tau f M /2}$, where $\tau$ is the membrane's time constant~\citep{Burkitt2006a}.

The number of selected afferents $M$ depends on the strategy~$n$. The probability that an afferent fires $k$ times in the $\Delta t$ window is given by the Poisson probability mass function: $\mathrm{P}(k\textrm{ spikes}) = \frac{\lambda^ke^{-\lambda}}{k!}$, with $\lambda=f\Delta t$. The probability that an afferent fires at least $n$ times is thus $1- e^{-\lambda} \sum\limits_{k=0}^{n-1} \frac{\lambda^k}{k!}$, and finally, on average:
\begin{equation}
M = N \left( 1- e^{-\lambda} \sum\limits_{k=0}^{n-1} \frac{\lambda^k}{k!}\right).
\end{equation}

We now need to estimate $V_{\mathrm{max}}$. Intuitively, during the $\Delta t$ window, the effective input spike rate, which we call $r$, is typically higher than $fM$, because we deliberately chose the most active afferents. For example, using Strategy~$\#1$ with $\Delta t = 10$ ms ensures that this rate is at least 100Hz per afferent, even if $f$ is only a few Hz. More formally, Strategy~$\#n$ discards the afferents that emit fewer than $n$ spikes. This means on average the number of discarded spikes is
$N e^{-\lambda} \sum\limits_{k=0}^{n-1} \frac{k\lambda^k}{k!}
=N e^{-\lambda} \sum\limits_{k=1}^{n-1} \frac{\lambda^k}{(k-1)!}
  =N e^{-\lambda} \lambda \sum\limits_{k=1}^{n-1} \frac{\lambda^{k-1}}{(k-1)!}
=N e^{-\lambda} \lambda \sum\limits_{k=0}^{n-2} \frac{\lambda^k}{k!}$. Thus on average:
\begin{equation}
\begin{aligned}
r&=N/\Delta t \left( \lambda - e^{-\lambda} \lambda \sum\limits_{k=0}^{n-2} \frac{\lambda^k}{k!}  \right)  \\
&=Nf \left( 1 - e^{-\lambda} \sum\limits_{k=0}^{n-2} \frac{\lambda^k}{k!}  \right).
\end{aligned}
\end{equation}

We note $\overline{V}^{\infty}=\tau r$ the mean potential of the steady regime that would be reached if $\Delta t$ was infinite. We now want to compute the transient response. The LIF with instantaneous synapses and unitary synaptic weights obeys the following differential equation:
\begin{equation}
\label{eq:LIF}
\tau\frac{\dif{V}}{\dif{t}}=-V+\tau\sum\limits_{i}\delta(t-t_i),
\end{equation}
\noindent where $t_i$ are the presynaptic spike times. We first make the approximation of continuity, and replace the sum of Diracs by an equivalent firing rate $R(t)$:
\begin{equation}
\tau\frac{\dif{V}}{\dif{t}}=-V+\tau R(t).
\end{equation}
$R(t)$ should be computed on a time bin which is much smaller than $\tau$, but yet contains many spikes, to avoid discretization effects. In other words, this approximation of continuity is only valid for a large number of spikes in the integration window, that is if $r\tau>>1$, which for Strategy~\#1 leads to $Nf\tau>>1$.

Note that $R(t)=fM$ during the noise period, and $R(t)=r$ during the $\Delta t$ window (in the absence of jitter).

At this point it is  convenient to introduce the reduced variable $v(t)=\frac{V(t)-\overline{V}_{\mathrm{noise}}}{\overline{V}^{\infty}-\overline{V}_{\mathrm{noise}}}$, which obeys the following differential equation:

\begin{equation}
\label{eq:lif}
\tau\frac{\dif{v}}{\dif{t}}=-v+i(t),
\end{equation}

\begin{figure}[H]

\centering\includegraphics[width=0.9\linewidth]{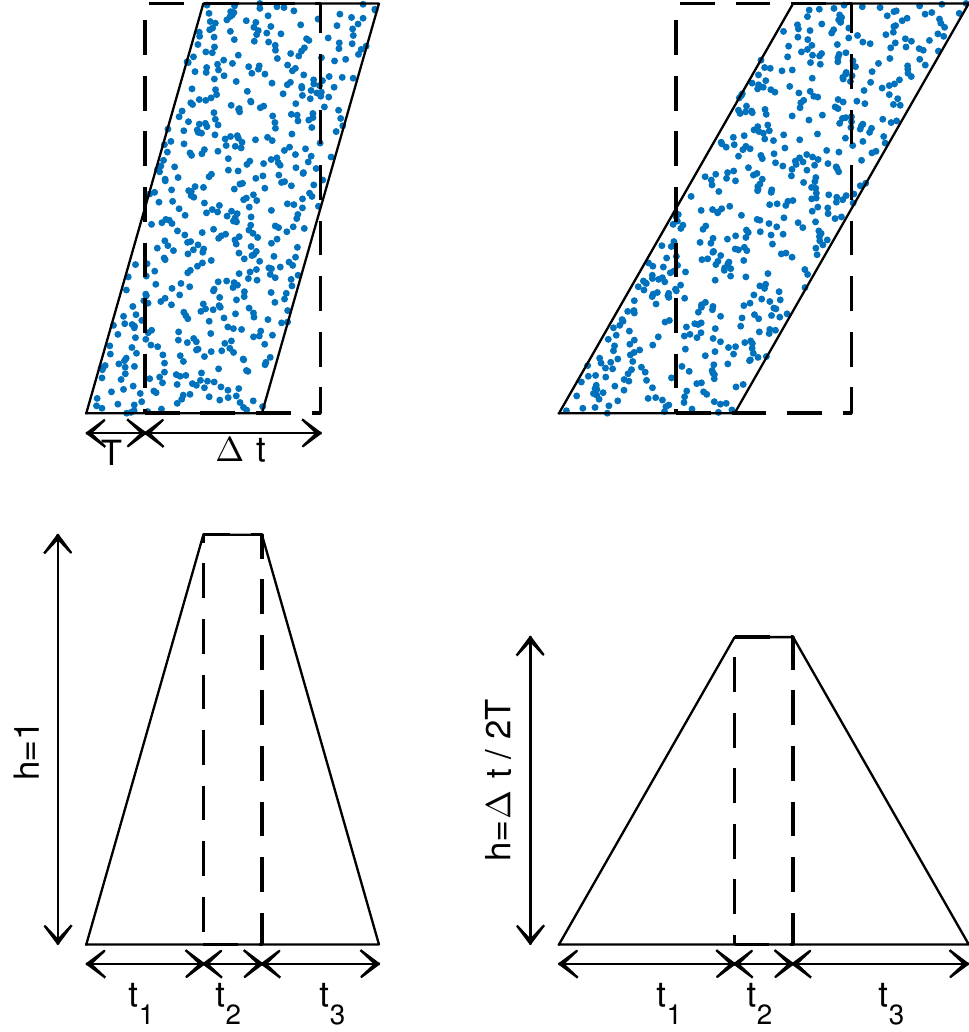}
\caption{Jittering the spike pattern. (Top) Raster plots for the $M$ selected afferents. x-axis is time, and y-axis is spike number (arbitrary, so we order them in increasing added jitter, which is a random variable uniformly distributed over $[-T,T]$). Dashed (resp. solid) lines corresponds to the boundaries of the raster plot before (resp. after) adding jitter. The left (resp. right) panel illustrates the $\Delta t>2T$ case (resp. $\Delta t<2T$ ) (Bottom) We plotted the corresponding spike time histograms, or, equivalently, doing the approximation of continuity, $i(t)$.  One can easily compute $t_1=t_3=\min(\Delta t, 2T)$, $t_2=|\Delta t-2T|$, and $h=\min(1,\Delta t/2T)$. One can check that the trapezoidal area is $\Delta t$ whatever $T$ (jittering does not add nor remove spikes).}
\label{fig:density}
\end{figure}

\noindent where $i(t)=\frac{R(t)-fM}{r-fM}$ is the dimensionless input current, such as $i=0$ during the noise period (when the input spike rate is $fM$), and $i=1$ when the input spike rate is $r$).

Without jitter, $i(t)$ would be a simple step function, equals to 1 during the $\Delta t$ window, and 0 elsewhere. Adding jitter, however, turns $i(t)$ into a trapezoidal function, which can be calculated (see Fig.~\ref{fig:density}). Now that $i(t)$ is known, one can compute $v(t)$ by integrating Equation~\ref{eq:lif}.

The response of the LIF to an arbitrary current $i(t)$ is~\citep{Tuckwell1988}:
\begin{equation}
v(t) = v_0e^{-t/\tau}+1/\tau\int_{0}^{t} e^{-(t-s)/\tau} i(s) \dif{s}.
\end{equation}
With $i=at+b$, and given that a primitive of $te^{t}$ is $te^{t}-e^{t}$,  the integral can be computed exactly:

\begin{equation}
\label{eq:v}
v(t)=a+b(t-\tau)+(v_0-a+b\tau)e^{-t/\tau}.
\end{equation}

Note that another jitter distribution than uniform (e.g. normal), would not lead to a piece-wise linear function for $i(t)$, and thus would typically not permit exact integration like here.

\begin{figure}[H]
\centering\includegraphics[width=0.9\linewidth]{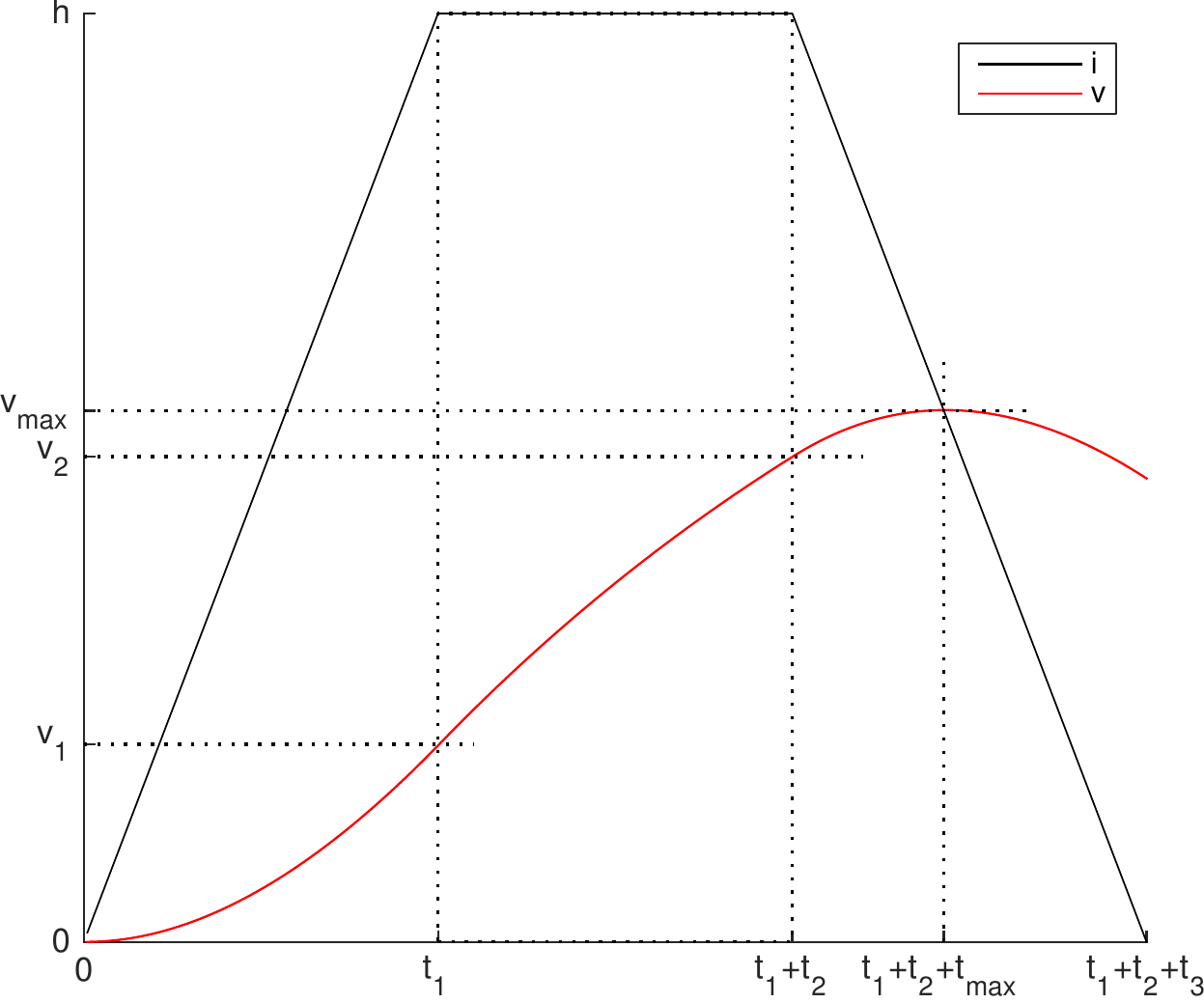}
\caption{$i(t)$ is piece-wise linear. $v(t)$, which lowpass filters $i(t)$, can be computed exactly on each piece. One can thus compute successively $v_1=v(t_1)$, $v_2=v(t_1+t_2)$, $v(t)$ for $t_1+t_2<t<t_1+t_2+t_3$ and its maximum $v_{\mathrm{max}}$, reached for $t=t_1+t_2+t_{\mathrm{max}}$.
}
\label{fig:v}
\end{figure}

As illustrated on Figure~\ref{fig:v}, one can use Equation \ref{eq:v} to compute successively $v_1=v(t_1)$, $v_2=v(t_1+t_2)$:

\begin{equation}
v_1=\frac{t_1+\tau(e^{-t_1/\tau}-1)}{2T},
\end{equation}

\begin{equation}
v_2=h+(v_1-h)e^{-t_2/\tau}.
\end{equation}

One can now compute $v(t)$ for $t_1+t_2<t<t_1+t_2+t_3$:
\begin{equation}
\label{eq:v3}
v(t+t_1+t_2)=h-\frac{t-\tau}{2T}+\left(v_2-h-\frac{\tau}{2T}\right)e^{-t/\tau},
\end{equation}
and differentiate it:
\begin{equation}
\frac{\dif{v(t+t_1+t_2)}}{\dif{t}}=-\frac{1}{2T}+\left(\frac{1}{2T}-\frac{v_2-h}{\tau}\right)e^{-t/\tau}.
\end{equation}
This derivative is 0, indicating that $v$ is maximal, for
\begin{equation}
\label{eq:tmax}
t_{\mathrm{max}}=t_1+t_2+\tau\log{\left(1+2T\frac{h-v_2}{\tau}\right)}.
\end{equation}
One can check that $v_{\mathrm{max}}=h-\frac{t_{\mathrm{max}}}{2T}$ which means that the maximum is on the trapezoid edge, which is logical: before the crossing $i>v$, so $v$ increases; after the crossing $i<v$, so $v$ decreases.
Plugging the $t_{max}$ value into Equation \ref{eq:v3}, and expliciting all variables, we have:
\begin{equation}
\begin{aligned}
&v_{\mathrm{max}} = \min \left( 1,\frac{\Delta t}{2T} \right) \\
&-\frac{\tau}{2T}\log \left(  1- e^{-\max(\Delta t, 2T)/\tau} + e^{-|\Delta t-2T|/\tau} \right).
\end{aligned}
\end{equation}

One can check that if $T<<\tau$ and $T<<\Delta t$, then $v_{\mathrm{max}}\sim1-e^{-\Delta t/\tau}$, which is the classical response of a LIF to a step current.

From the definition of $v$: $V_{\mathrm{max}}-\overline{V}_{\mathrm{noise}}=v_{\mathrm{max}}(V^{\infty}-\overline{V}_{\mathrm{noise}})$.
We now have everything we need to compute the signal to noise ratio:

\begin{equation}
\label{eq:SNR}
\begin{aligned}
&SNR = v_{\mathrm{max}} \frac{V^{\infty}-\overline{V}_{\mathrm{noise}}} {\sigma_{\mathrm{noise}}}\\
&= v_{\mathrm{max}} e^{-\lambda} \frac{\lambda^{n-1}}{(n-1)!}  \sqrt{\frac{2\tau N f}{1-e^{-\lambda} \sum\limits_{k=0}^{n-1} \frac{\lambda^k}{k!} }}.
\end{aligned}
\end{equation}

\subsection{Numerical validation}

\begin{figure}[ht]
\centering\includegraphics[width=0.6\linewidth]{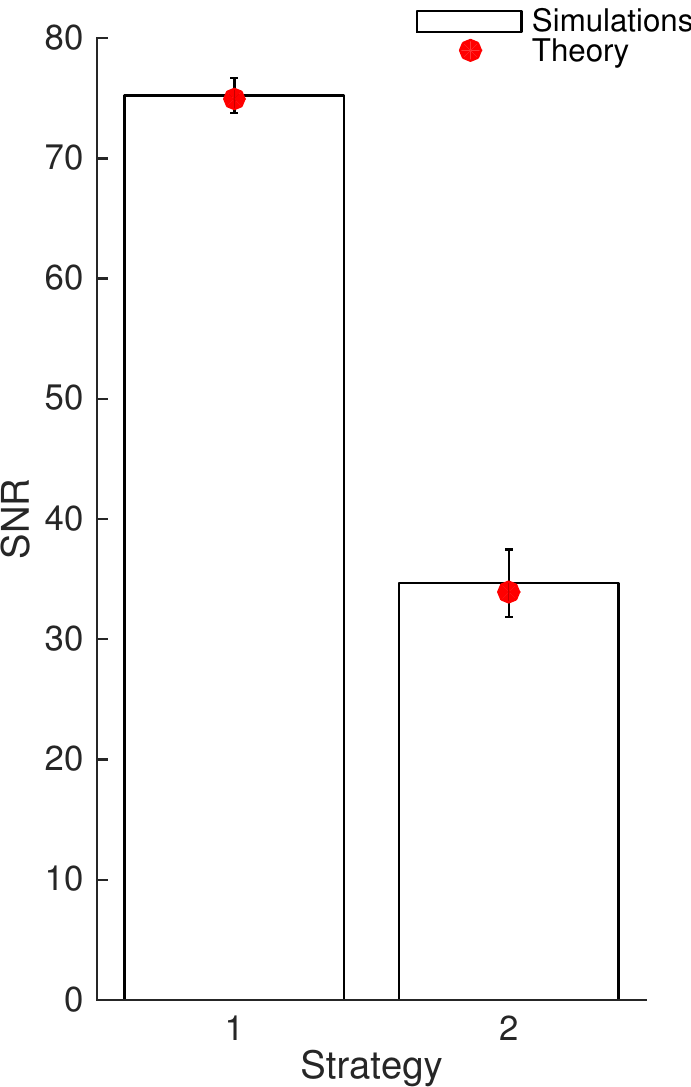}
\caption{Numerical validation of the theoretical SNR values, for strategies 1 and 2. Error bars show $\pm$1 s.d.
}
\label{fig:numerical}
\end{figure}

We verified the exact Equation~\ref{eq:SNR} through numerical simulations. We used a clock-based approach, and integrated Equation~\ref{eq:LIF} using the forward Euler method with a 0.1ms time bin. We generated 100 random Poisson patterns of duration $L=20$ms, involving $N=10^4$ neurons with rate $f=5$Hz. We chose
$\Delta t=L=20$ms, i.e. the LIF was connected to all the afferents that emitted at least $n$ spikes during the whole pattern, $n$ being the strategy number. In order to estimate $V_{\mathrm{max}}$, each pattern was presented 1000 times, every 400ms. Between pattern presentations, the afferents fired according to a Poisson process, still with rate $f=5$Hz, which allowed to estimate $\overline{V}_{\mathrm{noise}}$ and $\sigma_{\mathrm{noise}}$. We could thus compute the SNR from Equation~\ref{eq:snr_def} (and its standard deviation across the 100 patterns), which, as can be seen on Figure~\ref{fig:numerical}, matches very well the theoretical values, for strategies 1 and 2.

\subsection{Optimizing the SNR}

\begin{figure*}[ht]
\centering\includegraphics[width=.75\linewidth]{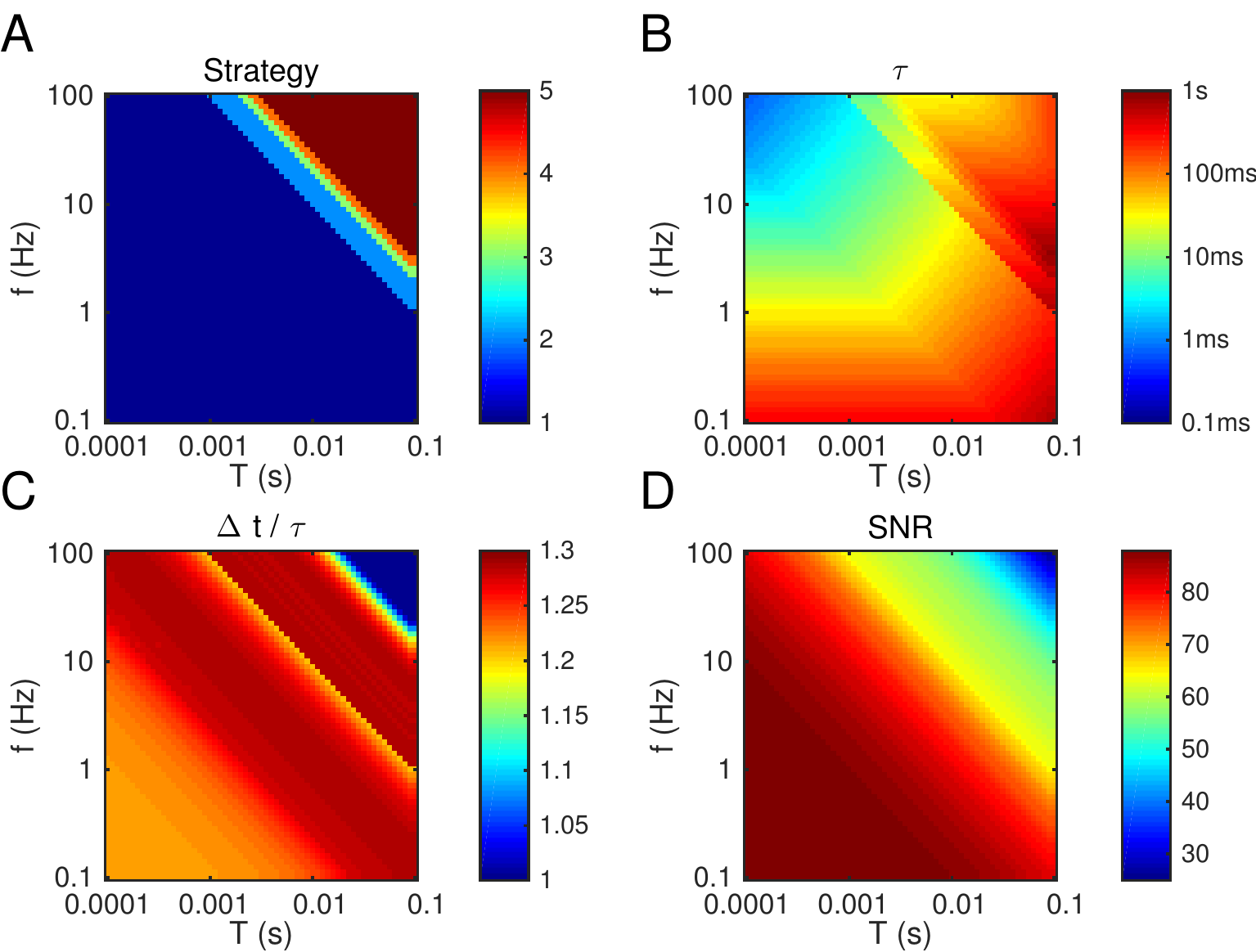}
\caption{Optimal parameters, as a function of $f$ and $T$. (A) Optimal strategy.  For clarity we only computed strategies 1..5, but it is clear that higher numbers would be optimal for large $f$ and $T$. (B) Optimal $\tau$ (note the logarithmic colormap). (C) Optimal $\Delta t$, divided by $\tau$. (D) Resulting SNR.
}
\label{fig:opt}
\end{figure*}

We now want to optimize the SNR given by Equation~\ref{eq:SNR}. We consider that $f$ and $T$ are external variables, and that we have the freedom to choose the strategy number $n$, $\tau$ and $\Delta t$. We also add the constraint $\tau fM\geq 10$, so that the approximation of continuity is reasonable, even in the noise periods. We assume that $L$ is sufficiently large so that an upper bound for $\Delta t$ is not needed. We used the Matlab R2015b Optimization Toolbox (MathWorks Inc., Natick, MA, USA) to compute the optimum numerically.

Figure~\ref{fig:opt} illustrates the results. One can make the following observations:
\begin{itemize}
\item Strategy \#1 is usually the best for $f$ and $T$ in the biological ranges (see below), while higher numbers are optimal for very large $f$ and $T$ (see panel A). This means that emitting a single spike is already a significant event, that should not be ignored. We will come back to this point in the discussion.
\item Unsurprisingly, optimal $\tau$  and $\Delta t$ typically have the same order of magnitude ($\Delta t$ being slightly larger, see panel C). Unless $T$ is high ($>$10ms), or $f$ is low ($<$1Hz), then these timescales should be relatively small (at most a few tens of ms). This means that even a long pattern (hundreds of ms or above) is optimally detected by a coincidence detector working at a shorter timescale. This could explain the apparent paradox between typical ecological stimulus durations (hundreds of ms or above) and the neuronal integration timescales (at most a few tens of ms). 
\item The constraint $\tau f M\geq 10$ imposes larger $\tau$ when both $f$ and $T$ are small (panel B, lower left). In the other cases, it is naturally satisfied.
\item Unsurprisingly, the optimal SNR decreases with $T$. What is more surprising, is that it also decreases with $f$. In other words, sparse activity is preferable. We will come back to this point in the discussion.
\end{itemize}

What is the biological range for $f$ and $T$? It is worth mentioning that $f$ is probably largely overestimated in the electrophysiological literature, because the technique totally ignores the cells that do not fire. Furthermore, experimentalists tend to select the most responsive cells, and search for stimuli that elicit strong responses. Mean firing rates, averaged across time and cells, could be smaller than 1 Hz~\citep{Shoham2006}.

$T$ corresponds to the spike time precision. Millisecond precision in cortex has been reported~\citep{Kayser2010,Panzeri2010,Havenith2011}. We are aware that other studies found poorer precision, but this could be due to uncontrolled variable or the use of inappropriate reference times~\citep{Masquelier2013}.

We now focus, as an example, on the point on the middle of the $T\times f$ plane, whose parameters are gathered in Table~\ref{tab:param}. The resulting SNR is very high (about 80). In other words, it is possible to choose a threshold for the LIF which will be reached when the pattern is presented, but hardly ever in the noise periods.

In the next section, we investigated, through numerical simulations, if STDP can find this optimum. More specifically, since STDP does not adjust $\tau$, we set it to the optimal value in Table~\ref{tab:param} and investigated whether STDP could lead to the optimal $n$ and $\Delta t$.

\begin{table}[ht]
\centering
\caption{Numerical parameters. First two lines correspond to external parameters, the rest of them are parameters to optimize.}
\begin{tabular}{l l }
\hline
\textbf{Parameter} & \textbf{Value} \\
\hline
$T$ & 3.2ms  \\
$f$ & 3.2Hz  \\
\hline
Optimal $\tau$ & 18ms  \\
Optimal $\Delta t$ & 23ms  \\
Optimal $n$ & 1  \\
\hline
\end{tabular}
\label{tab:param}
\end{table}

\section{Simulations show that STDP can be close-to-optimal}
\subsection{Set-up}

  The set up we used was similar to the one of our previous studies~\citep{Masquelier2008,Gilson2011}. We simulated a LIF neuron connected to all of the $N=10^4$ afferents with plastic synaptic weights $w_i\in[0,1]$, obeying the following differential equation:
 \begin{equation}
\label{eq:LIF_w}
\tau\frac{\dif{V}}{\dif{t}}=-V+\tau\sum\limits_{i,j}w_i(t_{ij})\delta(t-t_{ij}),
\end{equation}
 
 Initial synaptic weights were all equal. Then these synaptic weights evolved in $[0,1]$ with additive, all-to-all spike STDP like in~\cite{Song2000}. Yet we only modeled the Long Term Potentiation part of STDP, ignoring its Long Term Depression (LTD) term. Here LTD was modeled by a simple homeostatic term $w^{\mathrm{out}}<0$, which is added to each synaptic weight at each postsynaptic spike~\citep{Kempter1999}. Note that using a spike-timing-dependent LTD, could also lead to the detection of a repeating pattern, as demonstrated in  our earlier studies~\citep{Masquelier2008,Masquelier2009}, but less robustly, because it is more difficult to depress the synapses corresponding to afferents that do not spike in the repeating pattern.

As in~\cite{Song2000}, at each synapse $i$, we introduce the trace of presynaptic spikes $A_{\mathrm{pre}}^i$, which obeys the following differential equation:
\begin{equation}
\label{eq:A}
\tau_{\mathrm{pre}} \frac{\dif A_{\mathrm{pre}}^i}{\dif t} = - A_{\mathrm{pre}}^i.
\end{equation}
Furthermore:
\begin{itemize}
\item At each presynaptic spike: $A_{\mathrm{pre}}^i \rightarrow A_{\mathrm{pre}}^i+\delta A_{\mathrm{pre}}$.
\item At each postsynaptic spike: $w^i \rightarrow w^i+A_{\mathrm{pre}}^i+w^{\mathrm{out}}$ for $i=1..N$, then the weights are clipped in [0,1].
\end{itemize}

\noindent We used $\delta A_{\mathrm{pre}}=0.01$ and $\tau_{\mathrm{pre}}=20$ms, while $w^{\mathrm{out}}$ and the LIF threshold $\theta$ were systematically varied (see below). The refractory period was ignored for simplicity.

We used a clock-based approach, and integrated Equations~\ref{eq:LIF_w} and \ref{eq:A}  using the forward Euler method with a 0.1ms time bin. The Matlab code for these simulations will be made available in ModelDB~\citep{Hines2004}  once this paper is accepted in a peer-reviewed journal.

We now describe the way the input spikes were generated. Between pattern presentations, the input spikes were generated randomly with a homogeneous Poisson process with rate $f$ (see Table~\ref{tab:param}). The spike pattern with duration $L=100$ms was generated only once using the same Poisson process (frozen noise). The pattern presentations occurred every $400$ms (in previous studies, we demonstrated that irregular intervals did not matter~\citep{Masquelier2008,Gilson2011}, so here regular intervals were used for simplicity). At each pattern presentation, all the spike times were shifted independently by some random jitters uniformly distributed over $[-T,T]$ (see Table~\ref{tab:param}).

\subsection{Results: two optimal modes}

The theory developed in the previous sections ignored the LIF threshold (a difference of unconstrained potential was maximized). But in the simulations, one needs a threshold to have postsynaptic spikes, necessary for STDP. Since we did not know which threshold values $\theta$ could lead to the optimal $\Delta t$, we performed an exhaustive search over threshold values, using a geometric progression with a 1.1 ratio. Note that (from Equation~\ref{eq:LIF_w}) the threshold $\theta$ can be interpreted as the number of synchronous presynaptic spikes needed to reach the threshold from the resting potential if these spikes arrive through maximally reinforced synapses ($w=1$).

We also used a geometric progression with a 1.1 ratio to search for $w^{\mathrm{out}}$. This parameter tunes the strength of the LTD relative to the LTP, and thus influences the number of reinforced synapses after convergence. For each $\theta \times w^{\mathrm{out}}$ point, 100 simulations were performed with different random patterns, and computed the proportion $p$ of ``optimal'' ones (see below for the definition).

\begin{figure*}[ht]
\centering\includegraphics[width=1.0\linewidth]{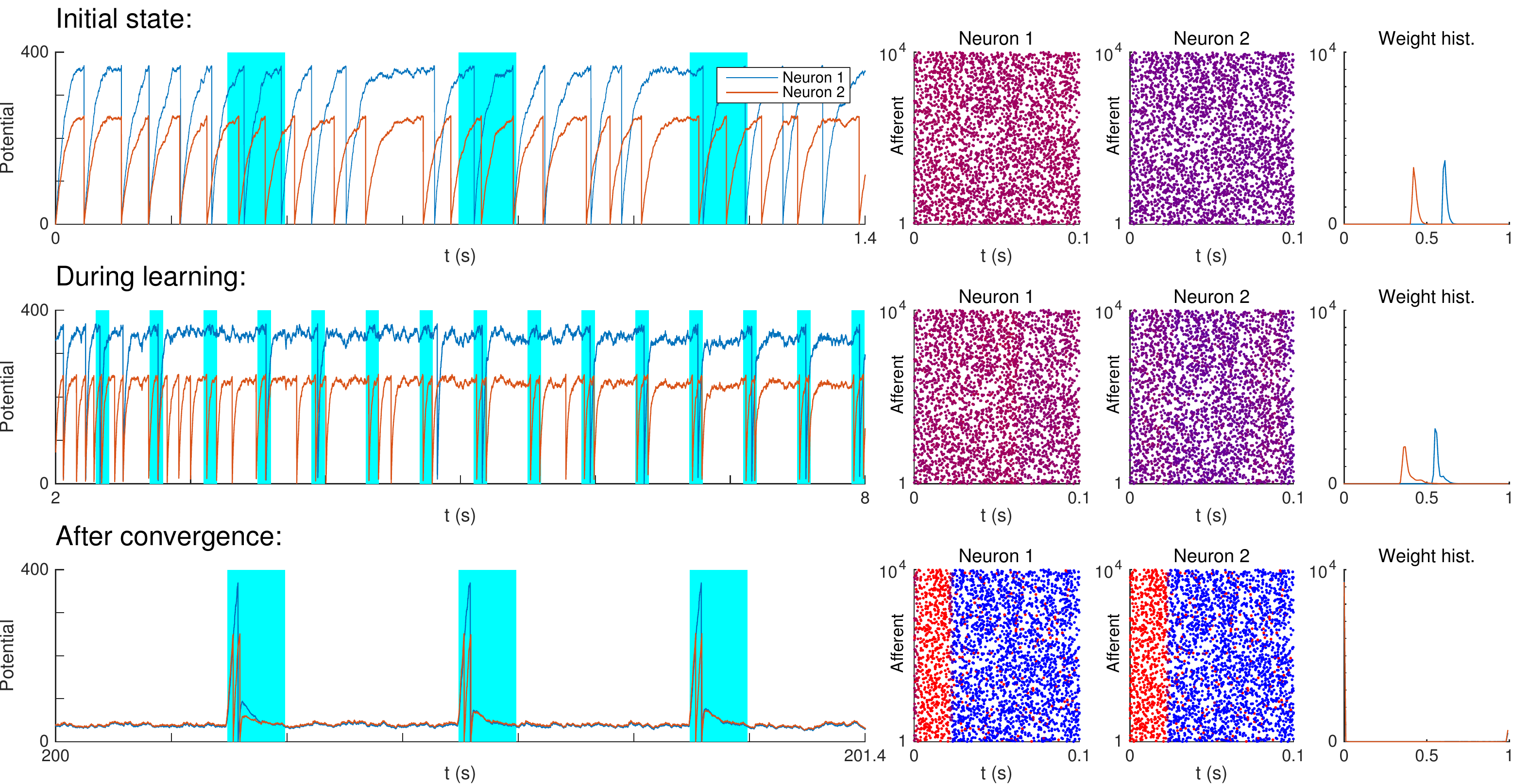}
\caption{ Unsupervised STDP-based pattern learning. Neuron \#1 and \#2 illustrate modes \#1 and \#2 respectively. (Top) Initial state. On the left, we plotted the potential of each neuron as a function of time. Cyan rectangles indicate pattern presentations. Next, we represented the weights  corresponding to the rightmost time point in two different ways. First, we plotted the  spike pattern, coloring the spikes as a function of the corresponding synaptic weight for each neuron: blue for low weight, purple for intermediate weight, and red for high weight. Initial weights were uniform (we used 0.68 for Neuron \#1 and 0.47 for Neuron \#2, in order to have $\overline{V}_{\mathrm{noise}}=\theta+2\sigma_{\mathrm{noise}}$). We also plotted the weight histogram for each neuron. (Middle) During learning. Selectivity emerges at $t\sim5$s, after $\sim$ 12 pattern presentations. Yet the weights still have intermediate values, leading to suboptimal SNR. (Bottom) After convergence. For both neurons, STDP has concentrated the weights on the afferents which fire at least once in a $\sim$ 23 ms long window, located at the beginning of the pattern. This results in 1 and 2 postsynaptic spikes for Neuron  \#1 and \#2 respectively each time the pattern is presented. Elsewhere both $\overline{V}_{\mathrm{noise}}$ and $\sigma_{\mathrm{noise}}$ are law, resulting in optimal SNR.
}
\label{fig:stdp}
\end{figure*}

The initial weights were computed such as $\overline{V}_{\mathrm{noise}}=\theta+2\sigma_{\mathrm{noise}}$ (leading to an initial firing rate of about 20Hz, see Fig.~\ref{fig:stdp} top). After 500 pattern presentations, the synaptic weights converged by saturation. That is synapses were either completely depressed ($w=0$), or maximally reinforced ($w=1$), as usual with additive STDP~\citep{Song2000,VanRossum2000,Gutig2003}. A simulation was considered optimal if the reinforced synapses did correspond to a set of afferents which fired at least once (Strategy \#1) in a subsection of the pattern, whose duration had to match the optimal $\Delta t$ window of the pattern given in Table~\ref{tab:param} (with a 10\% margin). In practice this subsection typically corresponded to the beginning of the pattern, because STDP tracks back through the pattern~\citep{Masquelier2008,Gilson2011}, but this is irrelevant here.

We found two optimal modes (see Fig.~\ref{fig:stdp}). The first one, with a high threshold ($\theta=370$) and strong LTD ($w^{\mathrm{out}}=-3.5\textrm{ }10^{-3}$) led to 1 postsynaptic spike at each pattern presentation (as in our previous studies~\citep{Masquelier2008,Masquelier2009,Gilson2011}). For this mode, $p=51$\%. The second mode, with a lower threshold ($\theta=250$) and weaker LTD ($w^{\mathrm{out}}=-1.6\textrm{ }10^{-3}$) led to 2 postsynaptic spikes at each pattern presentation, and $p=87$\% (the lower threshold increases the probability of false alarms during the noise period, but this problem could be solved by requiring two consecutive spikes for pattern detection). Figure~\ref{fig:stdp} illustrates an optimal simulation for both modes. We conclude that for most patterns, STDP can turn the LIF neuron into an optimal, or close-to-optimal pattern detector.

Detection is  optimal only after convergence (i.e. weight binarization), which takes time (about 500 pattern presentations). This is because the learning rate we used is weak ($\delta A_{\mathrm{pre}}=0.01$, in other words, the maximal weight increase caused by one pair of pre- and post-synaptic spike is only 1\% of the maximal weight), as in other theoretical studies and in accordance with experimental measurements~\citep{Song2000,Masquelier2008,Masquelier2009,Yger2015}. By using a higher rate, it is possible to converge faster, at the expense of the robustness. For example with $\delta A_{\mathrm{pre}}=0.02$, convergence occurs in $\sim$ 250 pattern presentations, but $p$ decreases to 44\% and 80\% for modes \#1 and \#2 respectively. In any case, it is worth mentioning that (suboptimal) selectivity emerges way before convergence (e.g. around  $t\sim5$s, or $\sim$ 12 pattern presentations in Figure~\ref{fig:stdp}).

Critically, for successful learning the pattern presentation rate must be high in the early phase of learning, before selectivity emerges. For example presenting the pattern every 800ms instead of 400ms leads to  $p=33\%$ and 43\% for modes \#1 and \#2 respectively. Once selectivity has emerged, this rate has much less impact, since the neuron tends to fire (and thus changes its weights) only at pattern presentations, whatever the intervals between them.

\section{Discussion}

 One of the main result of this study is that even a long pattern (hundreds of ms or above) is optimally detected by a coincidence detector working at a shorter timescale (tens of ms), and which thus ignores most of the pattern. One could have thought that using $\tau\sim L$, to integrate all the spikes from the pattern would be the best strategy. Instead, it is more optimal to use a subpattern as the signature for the whole pattern (see Fig.~\ref{fig:opt}).
  
  We also demonstrated that STDP can find the optimal signature in an unsupervised manner, by mere pattern repetitions. Note that the problem that STDP solves here is similar to the one addressed by the Tempotron \citep{Gutig2006}, which finds the best spike coincidence to separate two (classes of) patterns, by emitting or not a postsynaptic spike. Recently, the framework has been extended to fire more than one spike per pattern~\citep{Gutig2016}, like here (e.g. Neuron \#2 in Fig.~\ref{fig:stdp}). Yet these mechanisms require supervision.
  
  In this work we only considered single cell readout. But of course in the brain, it is likely that a population of cells is involved, and these cells could learn different subpatterns (lateral inhibition could encourage them to do so~\citep{Masquelier2009}). If each cell is selective to a subpart of the repeating pattern, how can one make a full pattern detector? One solution is to use one downstream neuron with appropriate delay lines~\citep{Carr1988}. Specifically, the conduction delays should compensate for the differences
of latencies, so that the downstream neuron receives the input spikes simultaneously if and only
if the sub-patterns are presented in the correct order. Another solution would be to convert the spatiotemporal firing pattern into a spatial one, using neuronal chains with delays as suggested by Tank and Hopfield~\citep{Tank1987}. Such a spatial pattern -- a set of simultaneously
  active neurons -- can then be learned by one downstream neuron equipped with STDP, and fully connected to the neuronal chains, as demonstrated in~\cite{Larson2010}. 

It   is also conceivable that the whole pattern is detected based on the mere number of subpattern detectors' spikes, ignoring their times. Two studies in the human auditory system are consistent with this idea: after  learning meaningless white noise sounds, recognition is still possible if the sounds are compressed or played backward~\citep{Agus2010}, or chopped into 10ms bins that are then played in random order~\citep{Viswanathan2016}.

    Our theoretical study suggests that synchrony is an important part of the neural code~\citep{Stanley2012}, that it is computationally efficient~\citep{Gutig2006,Brette2012}, and that coincidence detection is the main function of neurons~\citep{Abeles1982,Konig1996}. In line with this proposal, neurons \emph{in vivo} appear to be mainly fluctuation-driven, not mean-driven~\citep{Brette2012,Brette2015}. It remains unclear if other spike time aspects such as ranks~\citep{Thorpe1998} also matter.

Our results show that, somewhat surprisingly, lower firing rates lead to better signal-to-ratio. This could explain why average firing rates are so low in brain, possibly smaller than 1 Hz~\citep{Shoham2006}. It seems like neurons only fire when they need to signal an important event, and that every spike matters~\citep{Wolfe2010}.

\section*{Acknowledgments}
This research received funding from the European Research Council under the European Union’s $7^{th}$
Framework Program (FP/2007-2013) / ERC Grant Agreement n.323711 (M4 project). We thank Saeed Reza Kheradpisheh and Matthieu Gilson for the many insightful discussion we had about this work.

%\section{References}

%% The Appendices part is started with the command \appendix;
%% appendix sections are then done as normal sections
%% \appendix

%% \section{}
%% \label{}

%% References
%%
%% Following citation commands can be used in the body text:
%% Usage of \cite is as follows:
%%   \cite{key}          ==>>  [#]
%%   \cite[chap. 2]{key} ==>>  [#, chap. 2]
%%   \citet{key}         ==>>  Author [#]

%% References with bibTeX database:

%\bibliographystyle{model1-num-names}
%\bibliographystyle{model2-names.bst}\biboptions{authoryear}
%\singlespacing
\footnotesize
%\bibliographystyle{apa}
%\bibliography{../../library}

%% Authors are advised to submit their bibtex database files. They are
%% requested to list a bibtex style file in the manuscript if they do
%% not want to use model1-num-names.bst.

%% References without bibTeX database:

% \begin{thebibliography}{00}

%% \bibitem must have the following form:
%%   \bibitem{key}...
%%

% \bibitem{}

% \end{thebibliography}

\end{document}